# Mathematics, Recursion, and Universals in Human Languages

P. Gilkey, S. Lopez Ornat, and A. Karousou

**Abstract**: There are many scientific problems generated by the multiple and conflicting alternative definitions of linguistic recursion and human recursive processing that exist in the literature. The purpose of this article is to make available to the linguistic community the standard mathematical definition of recursion and to apply it to discuss linguistic recursion. As a byproduct, we obtain an insight into certain "soft universals" of human languages, which are related to cognitive constructs necessary to implement mathematical reasoning, i.e. mathematical model theory.[1]

1. **Introduction.**

The concept of recursion is central to the study of language. But the concept is often badly used because the definitions available in the literature are either wrong or are phrased ambiguously. Because recursion is a central notion in linguistics, it has to be made clear in the linguistic context. The notion of recursion is easy to define mathematically – and we shall say that a language is recursive if it is possible to implement the necessary mathematical concepts in the language under consideration. We believe that with this new formalism we are clarifying many foundational questions and that we are helping all researchers in fields in which linguistic recursion is relevant.

In this article, rather than trying to impose on human languages what we feel to be the artificial structure of mathematical logic, we shall explore similarities between some basic structures of mathematics and of human language. Our focus will be conceptual and we shall attempt to outline a broad framework rather than attempting either mathematical proofs or rigorous scientific experiments. We are interested in the extent to which it is possible to encode the language of mathematics in natural human languages. And here we use the words "mathematics" in a cognitive structural sense that can be made precise using a branch of mathematics called "mathematical model theory".

Many authors have drawn analogies previously between formal mathematical systems and human languages. Symbolic logic has been used to study linguistics – a task for which it is remarkably poorly suited. Working mathematicians quickly discover that what is important about symbolic logic is that it exists, i.e. that it is (in principle) possible to quantify mathematics in a very formal precise system – this fact is central to a branch of mathematics called model theory. However, mathematicians seldom or never use symbolic logic for communicating mathematics orally. Furthermore, conversations in a mathematics common room are often very different than the formal structures presented in mathematics papers.

We shall content ourselves for the most part with the basic mathematical operations that would be familiar to an undergraduate mathematics major. Since the formalism of mathematical model theory can encode mathematical recursion, any natural language that can implement these

---

concepts a-fortiori implements recursion. And we shall in fact use this as our definition of recursion.

The cognitive structures we shall be discussing were obviously not created to implement mathematics. Still, it is convenient to talk about mathematics in order to have a rigorous framework for identifying the structures involved, as one must be very precise. Negation, constants, variables, conjunction, quantification, and implication have obvious significance in communicating information and in forming part of the ``internal operating system''.  We emphasize that language is, after all, above all else about communication. And communication is both external (i.e. between two or more individuals) and, perhaps some would argue, also internal (as the operating system that enables us to communicate with ourselves).

Here is a brief outline to the paper. In Section 2, we shall introduce some standard mathematical formalism and discuss what the basic elements of mathematical statements are. This properly belongs to the field of mathematical model theory but we shall not go into great detail. This naturally leads us in Section 3 to a discussion of "soft linguistic universals" and we shall give a more precise definition there. But, roughly speaking, by "soft linguistic universals" we mean properties that are shared by most, but not necessarily all, human languages. These are common structures, procedures, and rules. This commonality is not necessarily genetically based but may arise from shared experience and general cogitative traits of the human mind, or from social structures as well. Work of Clark [5, 6, 7], Davidson [8], Defaene et al [11], Fitch, Hauser, and Chomsky [14], and others is discussed in this context.

In Section 4, we turn to mathematical recursion. Our treatment is, of course, standard but perhaps not familiar to all who study languages. The notion of infinity is central to recursion and we also discuss discrete infinity. In Section 5, we apply mathematical recursion to the linguistic setting. Work of Clark [7], Hauser, Chomsky, and Fitch [13, 14], Marcus [19], Pinker and Jackendoff [22], Tomasello [30], and others is discussed from this point of view. Our conclusions are summarized in Section 6.

We emphasize that very little of this is new. But we think it important to bring together the fields of mathematics and linguistics as both have much to offer to each other.

**2. Mathematical formalism**

**2.1. The notion of a limit.** We begin our discussion with an illustration from calculus. In what follows, we shall let f(x) be a real valued function of a real variable x; defining this notion precisely is in itself instructive but we shall omit such a discussion in the interests of brevity. Newton would have said:

**Definition 2.1.** To say that L is the limit of f as x tends to a means that if x is infinitesimally close to (but different from) a, then f(x) is infinitesimally close to L.

This definition, when formalized, leads to a branch of mathematics called nonstandard analysis, see, for example, the discussion in [9, 18, 24]. But as it stands, one may not verify if this definition is true or false for particular examples; it is not operationalizable. Instead, we replace Definition 2.1 by the following:

**Definition 2.2** To say that L is the limit of f as x tends to a means that given any positive number ε, then there exists a positive number δ so 0<|x-a|< δ implies |f(x)-L|< ε.

Any elementary treatise on analysis provides many examples using this definition to prove mathematical assertions – this is an operationally useful concept. Rather than belaboring the point, we refer, for example, to Ross [26] for further details and examples.

**2.3 Mathematical model theory.** The English language plays a central role in Definition 2.2. However, one can go yet one step further, and be even more very formal and remove this dependence and render it symbolically. The following Definition would be appropriate for a course in nonstandard analysis (see Robinson [24] or Kanovei et al [18]) – the point being to abstract the essential logical structure involved:

**Definition 2.3.** $\{L=\lim(a,f)\} \Leftrightarrow \{\{\forall \varepsilon \in R\} \wedge \{\varepsilon > 0\}\} \Rightarrow \{\{\{\exists \delta \in R\} \wedge \{\delta > 0\}\} \Rightarrow \{\{\{\forall x \in R\} \wedge \{0 < |x-a|\} \wedge \{|x-a| < \delta\}\} \Rightarrow \{|f(x) - L| < \varepsilon\}\}\}$.

Although Definition 2.3 has totally lost contact with the English Language, it would be recognizable to any working mathematician regardless of what human language/s s/he spoke. However, it must be admitted, in gaining universality and precision, Definition 2.3 has lost ease of understanding and takes much effort to deconstruct[2]. The study of mathematical statements and arguments from this point of view belongs to a field called mathematical model theory.

**2.4 Basic elements of mathematical statements.** The basic elements in Definition 2.3 and in common mathematical arguments can be outlined follows:
  (1) Constants. One can distinguish several subclasses – the following list is not exhaustive:
    (a) Primitives: 0 means "zero" and "R" means "the real numbers".
    (b) Relations: "<" is a binary relation; a<b means "a is less b".
    (c) Functions: "|a|" is a 1-argument function whose input is a real number and whose output is the absolute value of that real number; "lim" is a 2-argument function whose first entry is a real number and whose second entry is a real valued function; it is not defined for all a and f.
  (2) Variables: x, a, L, ε, f, and δ.
  (3) Quantifiers: ∀ means "for every" or "given any"; ∃ means "there exists" or "one can find".
  (4) Implication: ⇒ means "implies that" or "such that"; ⇐ reverses the implication and means "is implied by"; ⇔ means both implications, i.e. "if and only if" or "equivalently".
  (5) Negation ¬P means "not P", i.e. "the statement P is false".
  (6) Conjunctions: ∨ means "and" while ∧ means "or".

Now this is not a minimal list since, for example, the assertion $\{P \Rightarrow Q\}$ is logically equivalent to the assertion $\neg \{P \wedge \neg Q\}$ and $\{P \wedge Q\}$ is logically equivalent to $\neg \{\{\neg P\} \vee \{\neg Q\}\}$. Thus one can eliminate ⇒ and ∧ from the list of basic elements. And, of course, variables and substitution are closely related but they are not identical concepts.

We emphasize. What is important is that one can in principle write down such formulas and admissible methods of argument very formally. A set S of axioms (mathematical statements) is said to be "consistent" if there exists a model satisfying these axioms. This is why the field called "model theory". It is a fundamental result in the subject that only finite constraints pertain, i.e. an infinite set S of axioms is consistent if and only if every finite subset T of S is consistent. As the models for each subset T may be different, it is necessary to "piece together" these different models using an ultra-filter to construct a single coherent model for S. Note that there can be many models for a given set of axioms; choosing a "non-standard" model for the real numbers permits one to discuss infinitesimals and thereby make Definition 2.1 mathematically precise and rigorous. The formalism of Section 2.4 is essential in this endeavor.

---

[2] And there is no guarantee we have typed it correctly.

## 3. Soft linguistic universals

The term "universals" in the linguistic literature usually refers to traits that arise solely from the genetic endowment. That is a Chomskian notion. The notion of genetic determinism on language acquisition is now part of folk-psychology, and it is widely popular in the fields of linguistics and of language acquisition. As an effect, the expression "Language Universal" is implicitly understood as "Language genetically-determined trait". We can call that concept a *hard* language universals' concept. In this paper, we choose to use the expression "soft linguistic universals", implying there are "universals" of language because people all over the world have similar communicative jobs to get done, and similar cognitive and social tools with which to do them, as Tomasello [31] words it (citing work of Bates 1979). Those constrictions, together with some hard-wired genetic general cognitive factors, determine that many (not all) languages may come to share some structural traits. One might add to that, that language acquisition is also a well defined problem for all human babies, so that complex human neural nets may come to similar and stable solutions for it. In this context, generalized structural properties of languages wouldn't be a surprise, and the properties shared would be *soft* universals. Thus we hope to bypass entirely questions concerning universal grammar. Similarly, one could ask whether mathematical model theory is only a formal cultural development. This is more or less akin to the philosophical question: do mathematicians invent or discover mathematics - i.e. is mathematics an innate reality or a human construct. Do the objects of mathematics have an intrinsic existence, which is extrinsic to humanity?

We have constants, variables, quantifiers, implication, negation, and conjunctions comprising the admissible statements of mathematical model theory given in Section 2.4. Linguistically speaking words and phrases such as "Chicago", "Peter" "belongs to", "is the daughter of", "windy", or "city" are all constants. Thus the sentence "Chicago is a windy city and Emily is the daughter of Peter" could be rendered as:

$$\{\{Chicago \in Windy\} \wedge \{Chicago \in City\} \wedge \{Emily \in Daughter(Peter)\}\}.$$

Pronouns such as "he", "it", "she", and "they" clearly are variables and admit substitution. Pronouns such as "some" and "all" are not variables however; they are closely linked to quantifiers. On the other hand, some verbs are constants like "belongs to" or "is a member of the set" ($\in$). Furthermore, in Spanish, the verb *haber* often means "there exists" and is a quantifier. Thus exact linguistic correspondences are misleading and probably should be avoided as different languages can implement these fundamental logical constructs differently.

As noted above, the use of the word "universal" in the linguistic literature can be misleading. Fitch, Hauser, and Chomsky [14] make this point clearly:

> *"The putative absence of obvious recursion in one of these languages is no more relevant to the human ability to master recursion than the existence of three-vowel languages calls into doubt the human ability to master a five or ten-vowel language".*

We note in a similar vein that the existence of a body of mathematical literature in a natural language which uses logical reasoning and the axiomatic method is a sufficient demonstration of the implementability of mathematical model theory in that language and establishes that that particular language is recursive. But, of course, the evidence need not be written, there may be purely linguistic oral indicators and perhaps the implementability of mathematical model theory does not rely upon a written form of the language. This is a fascinating question, which must be left open for future investigation.

**3.1. Implementation**. If there exist languages where mathematical model theory cannot be implemented, this is an interesting feature that distinguishes such languages from other languages with which we are familiar. And the phenomenon is not purely modern; evidence of mathematical literature in cuneiform provides strong (although obviously not conclusive) evidence that Babylonian implements mathematical model theory (see, for example, the discussion in [1, 17, 20]). And clearly the existence of Euclid's Elements shows that the ancient Greek language implements mathematical reasoning of a high order.

There are modern questions worth pursuing. Research by Dehaene et al [11] provides a tantalizing insight into Amazonian Indigene cultures:

> *"Our results suggest that all humans share the intuition that numbers map onto space, but that culture specific experiences alter the form of this mapping – and that a logarithmic scale is characteristic of infants with a shift to a linear mapping occurring later in Western children."*

While this is not directly related to the question we have been posing, which deals with the cognitive concepts underlying mathematical reasoning, it is nevertheless very suggestive. And there are similar glimpses into geometry posed by Dehane et al [10]:

> *"Does geometry constitute a core set of intuitions present in all humans, regardless of their language or schooling?"*

Unfortunately, the question we are considering is both more fundamental and less subject to investigation. We wish to know not how numbers are encoded nor whether geometrical insight is present. We wish to know whether the basic logical structure with which we perceive mathematics is universal.

We hope the universals of García Calvo [15] can be explained in this way as can the universal semantic primes of Goddard and Wierzbicka [16]. For us, if Everett [12] holds up, it is an interesting comment that not all languages can implement the logical structures underlying mathematics. It is not, however, the disaster it seems to be for Wierzbicka and Chomsky, among others. Also note the comment (cited by Everett [12]) by Davidson [8] that

> *"The last stage in language development requires a leap, it introduces quantification, the concepts expressed by the words some and all. Once we advance to this stage, we have arrived at languages that match or begin to match our own in complexity."*

Note again the central role played by the quantifiers ($\forall, \exists$). We wish to clarify to formalists (the first author being such) that it is necessary to be coherent with and to understand that perhaps the only acceptable definition of a "linguistic universal" may not in fact be linguistic at all but rather may be a general cognitive definition which is exemplified by the logical constructs of mathematical model theory.

**3.2. Efficiency**. Concerning the efficiency of implementing mathematical constructs, Clark [6] argues:

> *"Embodied agents use bodily actions and environmental interventions to make the world a better place to think in. Where does language fit into this emerging picture of the embodied, ecologically efficient agent? One useful way to approach this question is to consider language itself as a cognition-enhancing animal built structure .... By materializing thought in words, we create structures that are themselves proper objects of perception, manipulation, and (further) thought."*

This suggests strongly that embodying basic logical processes (such as exemplified in mathematical logic) in language enables "perception, manipulation, and (further) thought". Clark [6] quotes Dehaene and coauthors as presenting

> *"a compelling model of precise mathematical thought that reserves a special role for internal representations of language-specific number words."*

We concur – the ability to encode logical mathematical reasoning at whatever level is an essential task of language. Clark also argues [5]:

> *"Public language … is a species of external artifact whose current adaptive value is partially constituted by its role in re-shaping the kinds of computational space that our biological brains must negotiate in order to solve certain types of problems, or to carry out certain complex projects. This computational role of language has been somewhat neglected (not un-noticed but not rigorously pursued either) …. Speech and text, we have seen greatly extend the problem-solving capacities of humankind. More profoundly, the practice of putting thoughts into words alters the nature of human experience."*

**3.3. The operating system**. The view of language as forming an integral part of the internal operating system is not, of course, the viewpoint taken by the "translation picture" that has been described succinctly by Clark [7]:

> *"Language works its magic by being understood, and understanding is in turn conceived as consisting wholly in something like translation into some other content-matching (or content exceeding) format. Such a view depicts language as a kind of high-level code that needs to be compiled or interpreted (in the computer science sense) to do its work".*

Clark [7] in fact takes a somewhat different point of view arguing:

> *"Language, however, occupies a wonderfully ambiguous position on an hybrid cognitive stage, since it seems to straddle the internal-external borderline itself, looking one moment like any other piece of the biological equipment, and the next like a peculiarly potent piece of external cognitive structure".*

Fortunately, settling this particular point is not relevant to the question we wish to pursue – but it is suggestive in that a language which cannot implement the elementary basic logical constructs of Section 2.4 is perhaps therefore defective in an essential way in that a central piece of the cognitive structure is missing.

**4. Mathematical Recursion**

**4.1. The Peano axioms**. Mathematical induction is taught in the first week of many $2^{nd}$ year courses in analysis. For example, one has following Ross [26]:

> *"We denote the set {1, 2, 3,..} of all natural numbers by N. Each natural number n has a successor, namely n+1. One has the **Peano** axioms for this set:*
> *N1) 1 belongs to the natural numbers.*
> *N2) If n is a natural number, then its successor n+ 1 is a natural number.*

*N3) 1 is not the successor of any natural number. In particular, 0 is not a natural number.*
*N4) If n and m are natural numbers which have the same successor, then n= m.*
*N5 A subset of the natural numbers which contains 1 and which contains n+1 whenever it contains n must be all of the natural numbers."*

Of course, this begs the question "what is a set" and that can be a crucial point in non-standard analysis. One can also express the Peano axioms symbolically – and with considerable loss of clarity – as follows:

**Definition 4.1. (Peano Axioms)**
- N1)  $\{1 \in N\}$.
- N2)  $\{\{n \in N\} \Rightarrow \{n+1 \in N\}\}$.
- N3)  $\neg\{\{\exists n \in N\} \land \{n+1=1\}\}$.
- N4)  $\{\{\{n \in N\} \land \{m \in N\} \land \{n+1= m+1\}\} \Rightarrow \{n= m\}\}$.
- N5)  $\{\{\{S \subset N\} \land \{1 \in S\} \land \{\{n \in S\} \Rightarrow \{n+1 \in S\}\}\} \Rightarrow \{S= N\}\}$.

The Peano axioms form the basis for mathematical induction and hence for recursion. For each $n \in N$, let $P(n)$ be a list of statements or propositions that may or may not be true. The principle of mathematical induction asserts that all these statements are true provided $P(1)$ is true and provided $P(n+1)$ is true whenever $P(n)$ is true.

**4.2. Recursive functions**. A function f can be defined recursively. For example one sets n!=1·2·...· n (read n factorial). The difficulty, of course, is in the "...". One knows intuitively what is meant by this expression. Clearly one just multiplies together the first n integers. But this is not a mathematical definition. Rather, recursively, one uses the Peano axioms to define $f(1) := 1$ and then recursively (or inductively) sets

$$f(n):= n \cdot f(n-1).$$

There is an extensive literature concerning recursive functions. Such functions (in principle) can be implemented on a computer. There exist non-recursively definable functions – the literature is extensive and we give just a few references to provide a flavor [3, 21, 25, 27].

**4.3. Infinity**. One can also give a precise definition to the meaning of ∞ (infinity) in this regard – the notion of "discrete infinity" in linguistics is perhaps considerably less precise. We will spare the reader more symbolic expressions and proceed using ordinary mathematical notation. We say that a sequence $s(n)$ of real numbers tends to ∞ if it increases without limit. More precisely:

**Definition 4.2**. A sequence $s(n)$ of real numbers tends to ∞ if given any real number K, there exists an integer N so that if $n>N$, then $s(n) >K$.

**5. Linguistic Recursion**

**5.1. Symbolism and recursive processing**. The question of symbolism also is fundamental. Are humans first of all symbolic, then subsequently linguistic, then recursive, and finally mathematical? Being first symbolic (young children are not fully symbolic) has some implications. One could argue that some of the basic principles of mathematical model theory are only symbolic. But recursion enters since a symbol obtains all the constancies across the differences in the instances it refers to. And of course the notion of substitution into variables enters. And being symbolic, as humans are, they can manipulate the symbols, as every other thing is manipulated This is, of course, exactly how computer programs act by manipulating themselves and by making no distinction between the data and the program. Thus at some moment, inner symbol manipulation becomes recursive, and at that moment, meta-symbols would be created. This is at the core – this is how one can regard recursion in human processing and/or in human learning procedures. And not as a definition of a procedure that is specified in our genes.

We suspect that recursive processing is slowly developed in humans as they grow and they mature – we do not know if it is "genetically inbuilt" in any sense. We also suspect that developing recursive procedures depends upon generalizing to all cognitive domains a procedure that has been "discovered" in one domain (which could be language) and that is then applied to others. We simply do not know if recursion is innate or not. We emphasize that the notion of recursive processing is subtly different from the notion of mathematical recursion (which is a cognitive construct) and it is necessary to avoid confusion in this regard.

**5.2. Recursion in language**. Recursion plays a crucial role in most analysis of language. Marcus [19] notes:

> *"Any recursive scheme must have a set of primitives, a way of combining those primitives to form new complex entities, a way of ensuring that the arrangement of the elements matters (for example so that 12 is not 21) or that 'the cat is on the map' is not 'the map is on the cat' and a way of allowing new complex entities to participate in the combinatorial process."*

Marcus [19] also remarks in speaking of different models of human cognition that:

> *"... each of these proposals turns out to implement the same machinery as the symbol-manipulation account of recursion. Each of these models includes a systematic difference between atomic and complex units, a way of combining these units to form new complex units, and a means by which new complex units may in turn serve as input to further combinations."*

This is in congruence with our point of view.

It is the ability of language to encode or operationalize in an efficient way certain modalities, which is universal. To greatly over simplify: we believe that languages have the soft universal linguistic possibility to encode (i.e. implement) the fundamental logical structures of mathematical model theory given in Section 2.4. In other words, these basic mathematical logical structures are reflected in language. Note that we are not speaking of numbers or geometry here but rather the basic logical structure of mathematics (i.e. mathematical model theory as discussed by Robinson [24]).

The process of language learning itself is recursive in this sense, as is complex language processing by adults. This relates to the production of discourse or to writing, for example. But Music processing can also be recursive, and social cognition, and visual object deconstruction, and perhaps much of conscious processing engaged in complex problem solving. This leads us again to a previous question of whether recursion is a domain-specific property of language, or rather a general cognitive process, which is materialized through language (among others) and, perhaps, also enhanced by it.

It is possible to define the notion of "mathematical truth" for formulas in mathematical model theory recursively [9]. When some humans access their own recursive processing, then, they can consciously do many new things, like mathematics, or linguistics, etc. By accessing their own implicit recursive procedures, they explicitly build, discover, and define recursion. The notion of "recursivity" with which we are working is closely related to, but subtly different from, that notion of recursion. Human implicit recursive processing can perhaps be expressed as something like: "do it again", "take now those new elements you obtained (however abstract and symbolic they already are) and apply to them the same logic you had used before, and see what comes out now". That is, of course, very similar to the mathematical notion of a mathematical recursive function.

**5.3. Symbolic communication, numbers, and discrete infinity**. Note that we are not asserting that language is just an implementation of mathematical model theory – far from it. The elements that we identify as corresponding to mathematical model theory are surely simply the basics; they are necessary underpinnings of but are far from the sufficient conditions for a successful human language. Anyone with even a mild ear for language can distinguish spoken Korean, Japanese, and Chinese without understanding a word of these languages simply from the sound of the language; this facility to distinguish between linguistic groups has also been demonstrated vary early in children. So it is clear that language has a structure, a rhythm and a poetry far removed from the logical underpinnings that we have identified here. But with that essential caveat, we believe that a careful discussion of some soft universal elements of language from a mathematic model theoretic viewpoint is likely to be a fruitful one. And we hope this approach will be a felicitous undertaking, which will avoid the sterile formalism of previous attempts to apply symbolic logic to linguistics. Tomasello [30] notes:

> *"First, and most importantly, human linguistic communication is symbolic … human symbols are aimed at the attentional and mental states of others ... human beings use their linguistic symbols together in patterned ways, and those patterns, known as linguistic constructions, take on meanings of their own – deriving partly from the meanings of the individual symbols, but, over time, at least partly from the pattern itself."*

The following quote from Clark [7] relates to work of Dehaene:

> *"Most of us can't form any clear image of, e.g., 98-ness (unlike, say, 2-ness). But we appreciate nonetheless that the number '98' names a unique quantity between 97 and 99".*

This is, of course, exactly how the natural numbers N are described. To every number (say 97), there is a successor (98) and another successor (99) so that 98 lie exactly in between 97 and 99. So what is being discussed is made precise using the Peano axioms – it is the very essence of recursion. Clark also inquires:

> *"What is going on when you think the thought that '98 is one more than 97?'"*

We emphasize – this is not the question we are studying – we are content to recognize that what is being implemented is recursion. But how it is implemented in our biological system is clearly a fundamental one. Also relevant is Pinker and Jackendoff [22]:

> *"More striking is the possibility that numbers themselves (beyond those that can be subitized) are parasitic on language – that they depend on learning the sequence of number words, the syntax of number phrases, or both".*

They also note:

> *"Recursion consists of embedding a constituent in a constituent of the same type – for example a relative clause inside a relative clause which automatically confirms the ability to do adlibitium."*

This is subtly different from other definitions of recursion. And one that is very hard to quantify. On the other hand, it is possible to measure whether or not the necessary logical elements are present in a given natural language to implement Definition 4.1 – and thus this measures the existence of a recursive capacity precisely. Again – we are not saying Definition 4.1 describes linguistic recursion – we are only saying that the ability to implement it is a proof that linguistic recursion is present.

The ability to implement Definition 4.2 is a proof that "discrete infinity" is present in a language. Hauser, Chomsky, and Fitch [13] state:

> "The computational mechanism of recursion is recently evolved and unique to our species ... only those mechanisms underlying FLN – particularly its capacity for discrete infinity – are uniquely human."

Clearly we need a careful definition of these quantities – and this can also be dealt with precisely using the material given above. Hauser, Chomsky, and Fitch [14] also state:

> "Recursion is agreed by most modern linguistics to be an indispensable core computational ability underlying syntax and thus language ... there are no unambiguous demonstrations of recursion in other human cognitive domains with the only clear examples (mathematical formulas, computer programming) being clearly dependent on language."

Mathematical recursion and computer science are intimately linked, of course, and as noted programming languages implement recursion (see [4] for further details). The question of recursion in non-humans or non-linguistically in humans is a fascinating subject – see, for example, the discussion in [23,28,29,32]

## 6. CONCLUSIONS

The two central points of this article may be summarized as follows.

**6.1. Recursion**. We have given a precise definition of recursion – and note that is inescapably linguistically based. A language is said to be recursive if it is possible to implement the language of mathematics as described above. Thus it is possible to test empirically whether or not a given language is recursive. And in particular to settle whether or not the language of the Piraha as described by Everet [12] is recursive and consequently to verify whether or not recursion is a quintessential and distinctive characteristic of human languages as some linguists might claim – we believe this is likely to be the case owing to the "adaptive value" of Clark.

The argument of Clark is central – language permits the implementation of recursion – and it is difficult (or impossible) to test for recursion absent language. We remain unconvinced that it is possible to give a precise definition of recursion for animals – language is central. And we remain unconvinced that in the absence of language that a human cognitive system can be said to be recursive or that a mind can be said to be recursive. Language is a crucial tool and absent language, recursion may well not be present. The human mind has the facility for recursion. But absent the tools provided by language, it might well not be able to exercise this facility. There are some fascinating hints that this is related to linguistic acquisition (see, for example, work of Barr [2]). And, as Everett would suggest, some languages may not be able to implement recursion.

**6.2. Universals**. Our other point is that the "soft universals" of language, which have been observed previously, may be exactly those necessary to implement the logical constructs of mathematical model theory. The elements of Section 2.4 (negation, variables, constants, conjunction, quantifiers, and implication) are soft universals of recursive languages precisely because implementing recursion is not possible absent these elements. Both natural language and mathematics are constructs of the same cognitive system; with language, the human cognitive system is recursive, and some soft universals of language are necessary prerequisites to ensure recursivity.

**Acknowledgments**: Research of the authors partially supported by Project DGI SEJ2007-67810 and MTM2006-01432.

PG: Institute of Theoretical Science, University of Oregon, Eugene Or 97403 USA.
Email: gilkey@darkwing.uoregon.edu.

SLO: Dpto. Psicologa Basica II Facultad de Psicolog``iía Universidad Complutense de Madrid, Espania. Email: slornat@psi.ucm.es.

AK: Psychology Department, IST/University of Hertfordshire, Athens, Omirou 16, Paleo Falizo, Athans 17564 Greece. Email: helespa@otenetkarousou@ist.edu.gr.